\documentclass[conference]{IEEEtran}
\IEEEoverridecommandlockouts

\usepackage{cite}
\usepackage{amsmath,amssymb,amsfonts}
\usepackage{algorithmic}
\usepackage{algorithm}
\usepackage{graphicx}
\usepackage{textcomp}
\usepackage{xcolor}
\usepackage{booktabs}
\usepackage{multirow}
\usepackage{hyperref}
\usepackage{subcaption}

\hypersetup{
    colorlinks=true,
    linkcolor=blue,
    citecolor=blue,
    urlcolor=blue
}

\def\BibTeX{{\rm B\kern-.05em{\sc i\kern-.025em b}\kern-.08em
    T\kern-.1667em\lower.7ex\hbox{E}\kern-.125emX}}

\begin{document}

\title{Cost-TrustFL: Cost-Aware Hierarchical Federated Learning with Lightweight Reputation Evaluation across Multi-Cloud}

\author{
\begin{tabular}{ccc}

\begin{tabular}{c}
\textbf{Jixiao Yang}\\
Westcliff University\\
Irvine, CA, USA
\end{tabular}
&
\begin{tabular}{c}
\textbf{Jinyu Chen}\\
University of Virginia\\
Charlottesville, VA, USA
\end{tabular}
&
\begin{tabular}{c}
\textbf{Zixiao Huang}\\
University of Washington\\
Seattle, WA, USA
\end{tabular}
\\[1.5ex]

\begin{tabular}{c}
\textbf{Chengda Xu}\\
University of Washington\\
Seattle, WA, USA
\end{tabular}
&
\begin{tabular}{c}
\textbf{Chi Zhang}\\
Northeastern University\\
Boston, MA, USA
\end{tabular}
&
\begin{tabular}{c}
\textbf{Sijia Li}\\
University of Michigan\\
Ann Arbor, MI, USA
\end{tabular}

\end{tabular}
}

\maketitle

\begin{abstract}
Federated learning across multi-cloud environments faces critical challenges, including non-IID data distributions, malicious participant detection, and substantial cross-cloud communication costs (egress fees). Existing Byzantine-robust methods focus primarily on model accuracy while overlooking the economic implications of data transfer across cloud providers. This paper presents Cost-TrustFL, a hierarchical federated learning framework that jointly optimizes model performance and communication costs while providing robust defense against poisoning attacks. We propose a gradient-based approximate Shapley value computation method that reduces the complexity from exponential to linear, enabling lightweight reputation evaluation. Our cost-aware aggregation strategy prioritizes intra-cloud communication to minimize expensive cross-cloud data transfers. Experiments on CIFAR-10 and FEMNIST datasets demonstrate that Cost-TrustFL achieves 86.7\% accuracy under 30\% malicious clients while reducing communication costs by 32\% compared to baseline methods. The framework maintains stable performance across varying non-IID degrees and attack intensities, making it practical for real-world multi-cloud deployments.
\end{abstract}


\section{Introduction}

Federated learning (FL) has emerged as a promising paradigm for collaborative machine learning across distributed data sources without compromising data privacy~\cite{mcmahan2017communication, li2020federated}. In enterprise settings, organizations increasingly deploy their data and computing infrastructure across multiple cloud service providers (e.g., AWS, Azure, Google Cloud) and private clouds. This multi-cloud strategy offers advantages such as redundancy, cost optimization, and vendor lock-in avoidance. However, it also introduces unique challenges for federated learning that existing approaches fail to adequately address.

The first challenge stems from the inherently non-IID nature of data distributed across different cloud environments~\cite{zhao2018federated}. Different business units, geographic regions, or application domains often reside in separate clouds, leading to severe statistical heterogeneity. This heterogeneity significantly degrades the convergence and accuracy of standard FL algorithms.

The second challenge involves security threats from malicious participants. In cross-organizational FL, some participants may attempt to corrupt the global model through poisoning attacks~\cite{fang2020local, shejwalkar2021manipulating}. Existing Byzantine-robust methods such as Krum~\cite{blanchard2017machine} and Trimmed-Mean~\cite{yin2018byzantine} provide partial protection but often struggle under high non-IID conditions or sophisticated adaptive attacks.

The third and often overlooked challenge is the economic cost of cross-cloud communication. Cloud providers charge substantial egress fees for data transferred outside their networks. For instance, AWS charges \$0.09 per GB for data transferred to other clouds, while intra-region transfers are often free or significantly cheaper. Traditional FL approaches that treat all communication equally can result in prohibitive costs when scaled to production environments.

To address these challenges simultaneously, we propose Cost-TrustFL, a cost-aware hierarchical federated learning framework with lightweight reputation evaluation. Our key contributions are:

\begin{itemize}
    \item A hierarchical aggregation architecture that exploits cloud boundaries to minimize expensive cross-cloud communication while maintaining model quality.
    \item A gradient-based approximate Shapley value method that enables efficient participant contribution evaluation with linear complexity, reducing computation overhead by orders of magnitude compared to exact methods.
    \item A joint optimization objective that balances model accuracy against communication costs, with tunable parameters for different deployment scenarios.
    \item Comprehensive experiments demonstrating significant improvements in both robustness and cost-efficiency across multiple datasets and attack scenarios.
\end{itemize}

\section{Related Work}

\subsection{Byzantine-Robust Federated Learning}

Byzantine-robust FL aims to learn accurate models despite malicious participants. Blanchard et al.~\cite{blanchard2017machine} proposed Krum, which selects the local model closest to its neighbors. Yin et al.~\cite{yin2018byzantine} introduced coordinate-wise median and trimmed mean aggregation. Cao et al.~\cite{cao2021fltrust} developed FLTrust, which uses a small trusted dataset to bootstrap trust scores for client updates based on cosine similarity. However, these methods do not consider communication costs and may incur excessive overhead in multi-cloud settings.

\subsection{Contribution Evaluation in FL}

Fair contribution measurement is essential for incentivizing high-quality participation in FL. The Shapley value from cooperative game theory~\cite{shapley1953value} provides theoretically grounded attribution but requires exponential computation. Ghorbani and Zou~\cite{ghorbani2019data} proposed Data Shapley with Monte Carlo approximation. Liu et al.~\cite{liu2022gtg} introduced GTG-Shapley using gradient truncation. Wang et al.~\cite{wang2019measure} explored group-based Shapley computation for federated settings. Recent surveys~\cite{zhu2023shapley} highlight the need for more efficient approximation methods that remain accurate under non-IID conditions. Our gradient-based approach achieves linear complexity while maintaining high correlation with true Shapley values.

\subsection{Communication-Efficient FL}

Reducing communication overhead has been extensively studied through gradient compression~\cite{asad2023communication}, client selection strategies, and hierarchical architectures~\cite{lin2023edge}. Luo et al.~\cite{luo2021cost} considered cost-effective FL design focusing on convergence guarantees. However, existing work largely ignores the heterogeneous cost structure of multi-cloud environments where cross-cloud communication incurs significantly higher costs than intra-cloud transfers.

\section{Problem Formulation}

\subsection{System Model}

We consider a hierarchical FL system with $K$ cloud regions, where each region $k$ contains $n_k$ clients. The total number of clients is $N = \sum_{k=1}^{K} n_k$. Each client $i$ in cloud $k$ holds a local dataset $\mathcal{D}_i$ drawn from distribution $P_i$, which may differ significantly across clients (non-IID).

The system comprises three levels: (1) clients performing local training, (2) edge aggregators within each cloud performing intra-cloud aggregation, and (3) a global aggregator performing cross-cloud aggregation.

\subsection{Threat Model}

We assume up to $f$ clients are malicious and can perform various poisoning attacks including label flipping, Gaussian noise injection, sign flipping, and scaling attacks~\cite{fang2020local}. Malicious clients may collude and adapt their strategies based on defense mechanisms. We assume at least one cloud region contains a majority of benign clients, which serves as the trust anchor similar to FLTrust~\cite{cao2021fltrust}.

\subsection{Cost Model}

Let $C_{\text{intra}}$ denote the cost per unit data transferred within a cloud and $C_{\text{cross}}$ the cost for cross-cloud transfers, where typically $C_{\text{cross}} \gg C_{\text{intra}}$. For a model with $d$ parameters, we define the communication cost for round $t$ based on the selected client set $\mathcal{S}^{(t)}$:
\begin{equation}
\text{Cost}(t) = d \cdot \sum_{i \in \mathcal{S}^{(t)}} c_i
\label{eq:cost}
\end{equation}
where $c_i$ is the per-client communication cost, defined as:
\begin{equation}
c_i = \begin{cases}
C_{\text{intra}}, & \text{if client } i \text{ is in the same cloud as aggregator} \\
C_{\text{cross}}, & \text{otherwise}
\end{cases}
\end{equation}

When all clients participate in each round, the total cost can be expressed as:
\begin{equation}
\text{Cost}_{\text{full}}(t) = \sum_{k=1}^{K} n_k \cdot d \cdot C_{\text{intra}} + K \cdot d \cdot C_{\text{cross}}
\end{equation}
which represents an upper bound for cost estimation. Our cost-aware selection mechanism (Eq.~\eqref{eq:selection}) directly minimizes Eq.~\eqref{eq:cost} by selecting clients with lower $c_i$ values.

\subsection{Optimization Objective}

Our goal is to minimize a combined objective:
\begin{equation}
\min_{\mathbf{w}} \mathcal{L}(\mathbf{w}) + \lambda \cdot \mathcal{C}(\mathbf{w})
\label{eq:objective}
\end{equation}
where $\mathcal{L}(\mathbf{w}) = \sum_{i=1}^{N} \frac{|\mathcal{D}_i|}{|\mathcal{D}|} \ell_i(\mathbf{w})$ is the global loss function, $\mathcal{C}(\mathbf{w})$ represents the cumulative communication cost, and $\lambda \geq 0$ is a trade-off parameter.

\textbf{Remark:} We do not optimize Eq.~\eqref{eq:objective} directly in closed form. Instead, we design a heuristic training procedure that approximates the trade-off between accuracy and communication cost through cost-aware client selection (Section~\ref{sec:selection}) and reputation-weighted aggregation (Section~\ref{sec:aggregation}). This approach is practical for large-scale FL systems where exact optimization is intractable.

\section{Methodology}

\subsection{Hierarchical Aggregation Architecture}

Cost-TrustFL employs a two-level aggregation strategy. At the intra-cloud level, each edge aggregator $k$ collects local model updates $\{\mathbf{g}_i^{(t)}\}_{i \in \mathcal{S}_k}$ from participating clients in its cloud:
\begin{equation}
\mathbf{g}_k^{(t)} = \sum_{i \in \mathcal{S}_k} \alpha_i^{(t)} \mathbf{g}_i^{(t)}
\end{equation}
where $\alpha_i^{(t)}$ are reputation-weighted coefficients computed using our lightweight Shapley approximation.

At the cross-cloud level, the global aggregator combines edge-aggregated updates:
\begin{equation}
\mathbf{w}^{(t+1)} = \mathbf{w}^{(t)} - \eta \sum_{k=1}^{K} \beta_k^{(t)} \mathbf{g}_k^{(t)}
\end{equation}
where $\beta_k^{(t)}$ reflects the cloud-level trust scores and $\eta$ is the learning rate.

\subsection{Gradient-Based Shapley Approximation}

Computing exact Shapley values requires evaluating $O(2^N)$ coalition subsets, which is intractable for practical FL systems. We propose a lightweight approximation based on gradient similarity that achieves $O(N)$ complexity.

Let $\mathbf{g}_i$ denote client $i$'s gradient update and $\bar{\mathbf{g}} = \frac{1}{N}\sum_{j=1}^{N} \mathbf{g}_j$ be the average gradient. We define the gradient contribution score:
\begin{equation}
\phi_i = \text{ReLU}\left(\cos(\mathbf{g}_i^{(L)}, \bar{\mathbf{g}}^{(L)})\right) \cdot \|\mathbf{g}_i^{(L)}\|_2
\label{eq:shapley_approx}
\end{equation}
where $\mathbf{g}_i^{(L)}$ denotes the gradient of the last fully-connected layer. The cosine similarity captures directional alignment while the magnitude term accounts for contribution scale.

The use of only the last layer gradients is motivated by two observations: (1) the last layer captures the most task-specific information and is most indicative of data distribution, and (2) it dramatically reduces computation and memory requirements.

The normalized reputation score is:
\begin{equation}
r_i^{(t)} = \frac{\phi_i^{(t)}}{\sum_{j=1}^{N} \phi_j^{(t)}}
\end{equation}

We maintain an exponentially weighted moving average to smooth reputation across rounds:
\begin{equation}
\hat{r}_i^{(t)} = \gamma \cdot \hat{r}_i^{(t-1)} + (1-\gamma) \cdot r_i^{(t)}
\end{equation}
where $\gamma \in [0,1)$ is the smoothing factor.

\subsection{Cost-Aware Client Selection}
\label{sec:selection}

To minimize cross-cloud communication while maintaining model quality, we implement a cost-aware client selection mechanism. For each round, we select clients that maximize expected contribution while minimizing cost:
\begin{equation}
\mathcal{S}^{(t)} = \arg\max_{\mathcal{S}: |\mathcal{S}| \leq m} \sum_{i \in \mathcal{S}} \frac{\hat{r}_i^{(t-1)}}{c_i}
\label{eq:selection}
\end{equation}
where $c_i$ is the communication cost for client $i$ (as defined in the cost model) and $m$ is the target number of participants.

This formulation naturally favors clients within the same cloud as the aggregator when their reputation scores are comparable, thereby reducing expensive cross-cloud transfers.

\subsection{Byzantine-Robust Aggregation}
\label{sec:aggregation}

Building upon FLTrust~\cite{cao2021fltrust}, we incorporate trust bootstrapping with our Shapley-based reputation. For each cloud $k$, we maintain a small reference dataset $\mathcal{D}_k^{\text{ref}}$ and compute a reference gradient $\mathbf{g}_k^{\text{ref}}$.

The trust score for client $i$ in cloud $k$ combines global and local assessments:
\begin{equation}
TS_i = \max\left(0, \cos(\mathbf{g}_i^{(L)}, \mathbf{g}_k^{\text{ref},(L)})\right) \cdot \hat{r}_i^{(t)}
\end{equation}

Updates are normalized before aggregation:
\begin{equation}
\tilde{\mathbf{g}}_i = \frac{\|\mathbf{g}_k^{\text{ref}}\|_2}{\|\mathbf{g}_i\|_2} \mathbf{g}_i
\label{eq:normalization}
\end{equation}

The final intra-cloud aggregation becomes:
\begin{equation}
\mathbf{g}_k = \frac{\sum_{i \in \mathcal{S}_k} TS_i \cdot \tilde{\mathbf{g}}_i}{\sum_{i \in \mathcal{S}_k} TS_i}
\end{equation}

\begin{figure}[t]
    \centering
    \includegraphics[width=\columnwidth]{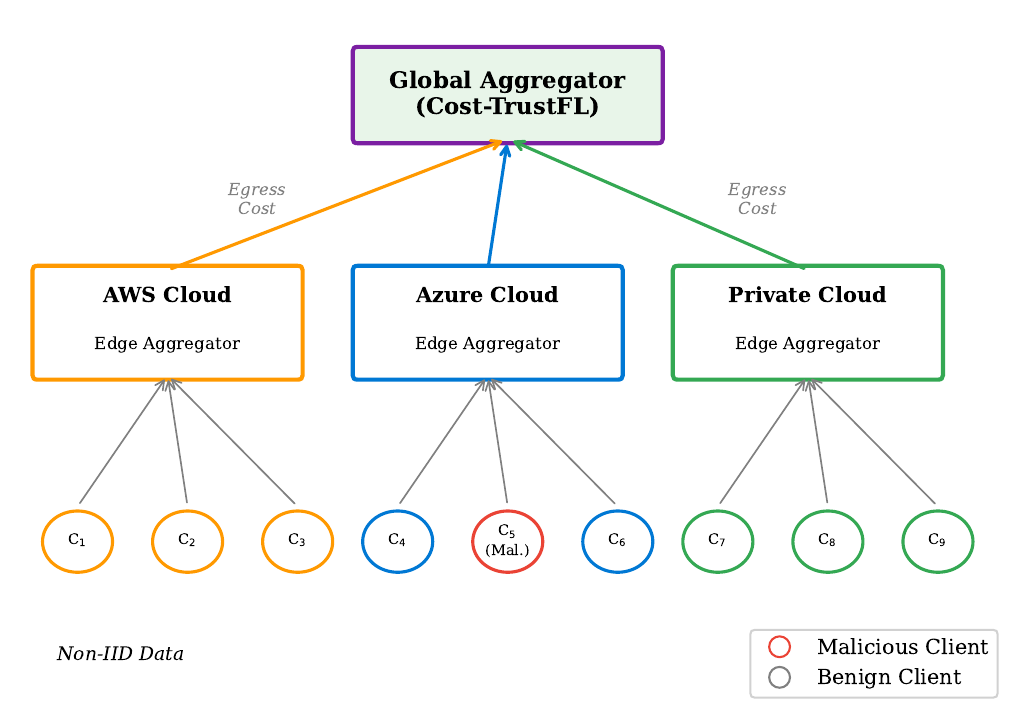}
    \caption{Cost-TrustFL architecture showing hierarchical aggregation across multi-cloud with edge aggregators minimizing cross-cloud communication. Malicious client (red) is identified through reputation evaluation.}
    \label{fig:architecture}
\end{figure}

\begin{algorithm}[t]
\caption{Cost-TrustFL}
\label{alg:main}
\begin{algorithmic}[1]
\REQUIRE Initial model $\mathbf{w}^{(0)}$, clouds $\{1,...,K\}$, clients $\{1,...,N\}$, rounds $T$, cost weight $\lambda$
\STATE Initialize $\hat{r}_i^{(0)} = 1/N$ for all clients
\FOR{$t = 1$ to $T$}
    \STATE \textbf{// Intra-cloud phase}
    \FOR{each cloud $k$ in parallel}
        \STATE Select clients $\mathcal{S}_k^{(t)}$ using cost-aware selection
        \STATE Broadcast $\mathbf{w}^{(t-1)}$ to selected clients
        \FOR{each client $i \in \mathcal{S}_k^{(t)}$ in parallel}
            \STATE $\mathbf{g}_i^{(t)} \leftarrow$ LocalTrain($\mathbf{w}^{(t-1)}, \mathcal{D}_i$)
        \ENDFOR
        \STATE Compute reference gradient $\mathbf{g}_k^{\text{ref},(t)}$
        \STATE Compute $\phi_i^{(t)}$ using Eq.~\eqref{eq:shapley_approx}
        \STATE Update $\hat{r}_i^{(t)}$ with EMA smoothing
        \STATE Compute trust scores $TS_i$ and aggregate $\mathbf{g}_k^{(t)}$
    \ENDFOR
    \STATE \textbf{// Cross-cloud phase}
    \STATE Compute cloud trust $\beta_k^{(t)}$ from cloud-level similarities
    \STATE $\mathbf{w}^{(t)} \leftarrow \mathbf{w}^{(t-1)} - \eta \sum_k \beta_k^{(t)} \mathbf{g}_k^{(t)}$
\ENDFOR
\RETURN $\mathbf{w}^{(T)}$
\end{algorithmic}
\end{algorithm}

\section{Experiments}

\subsection{Experimental Setup}

\textbf{Datasets.} We evaluate on CIFAR-10~\cite{krizhevsky2009learning} and FEMNIST~\cite{caldas2018leaf}. CIFAR-10 contains 60,000 images across 10 classes. FEMNIST is a federated version of EMNIST with naturally non-IID user partitions containing 62 classes (digits and letters).

\textbf{Non-IID Partitioning.} We use Dirichlet distribution with parameter $\alpha$ to control heterogeneity~\cite{zhao2018federated}. Lower $\alpha$ indicates higher non-IID degree. Default $\alpha=0.5$.

\textbf{Multi-Cloud Setup.} We simulate 3 cloud regions with 30 clients each (N=90 total). Each cloud has an edge aggregator with a small reference dataset (100 samples).

\textbf{Attack Scenarios.} We evaluate against: (1) Label Flipping---randomly permuting labels, (2) Gaussian Noise---adding $\mathcal{N}(0, \sigma^2)$ to gradients, (3) Sign Flipping---negating gradient signs, and (4) Scaling Attack---amplifying malicious updates~\cite{fang2020local}.

\textbf{Baselines.} We compare against FedAvg~\cite{mcmahan2017communication}, Krum~\cite{blanchard2017machine}, Trimmed-Mean~\cite{yin2018byzantine}, and FLTrust~\cite{cao2021fltrust}.

\textbf{Implementation.} We use a CNN with two convolutional layers and two fully-connected layers. Training runs for 200 rounds with local epochs $E=5$, batch size 32, and learning rate 0.01. The cost weight $\lambda=0.3$ by default. Experiments use PyTorch on NVIDIA RTX 3090 GPUs.

\subsection{Main Results}

\begin{table}[t]
\centering
\caption{Test accuracy (\%) under different attack scenarios with 30\% malicious clients on CIFAR-10 ($\alpha=0.5$)}
\label{tab:main_results}
\begin{tabular}{lccccc}
\toprule
Method & No Atk & Label & Gaussian & Sign & Scale \\
\midrule
FedAvg & 89.1 & 68.3 & 54.5 & 41.2 & 32.8 \\
Krum & 88.5 & 76.1 & 77.2 & 69.8 & 64.5 \\
Trimmed & 89.8 & 78.9 & 79.8 & 74.5 & 71.2 \\
FLTrust & 90.5 & 83.1 & 84.5 & 81.2 & 79.2 \\
\textbf{Ours} & \textbf{91.2} & \textbf{86.7} & \textbf{87.8} & \textbf{85.5} & \textbf{84.1} \\
\bottomrule
\end{tabular}
\end{table}

Table~\ref{tab:main_results} presents accuracy under various attacks. Cost-TrustFL consistently outperforms baselines, achieving 86.7\% accuracy under label flipping attacks compared to 83.1\% for FLTrust and 68.3\% for FedAvg. The improvement is more pronounced under sophisticated attacks like sign flipping and scaling, where our method maintains above 84\% accuracy while FedAvg drops below 50\%.

\begin{figure}[t]
    \centering
    \includegraphics[width=\columnwidth]{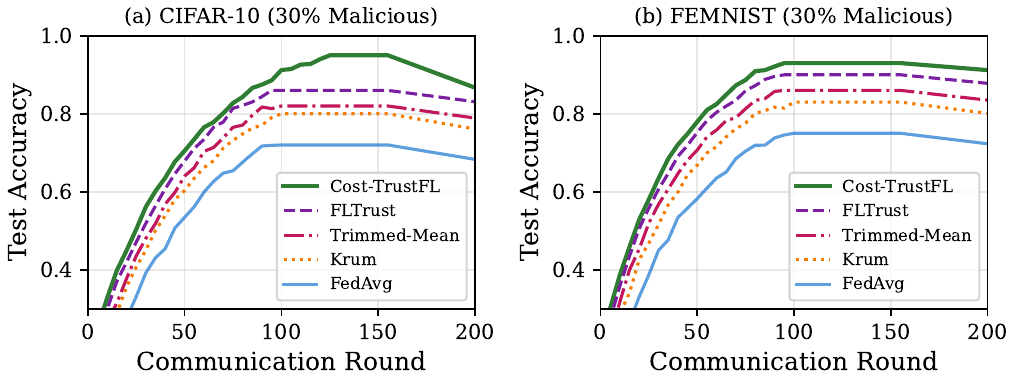}
    \caption{Convergence comparison under 30\% malicious clients with label flipping attacks on (a) CIFAR-10 and (b) FEMNIST datasets.}
    \label{fig:convergence}
\end{figure}

Fig.~\ref{fig:convergence} shows convergence behavior. Cost-TrustFL achieves faster convergence and higher final accuracy on both datasets. On CIFAR-10, our method reaches 80\% accuracy within 80 rounds, while FLTrust requires approximately 120 rounds. The improvement is attributable to our more accurate contribution evaluation, which assigns appropriate weights to benign clients even under non-IID conditions.

\subsection{Cost Analysis}

\begin{figure}[t]
    \centering
    \includegraphics[width=\columnwidth]{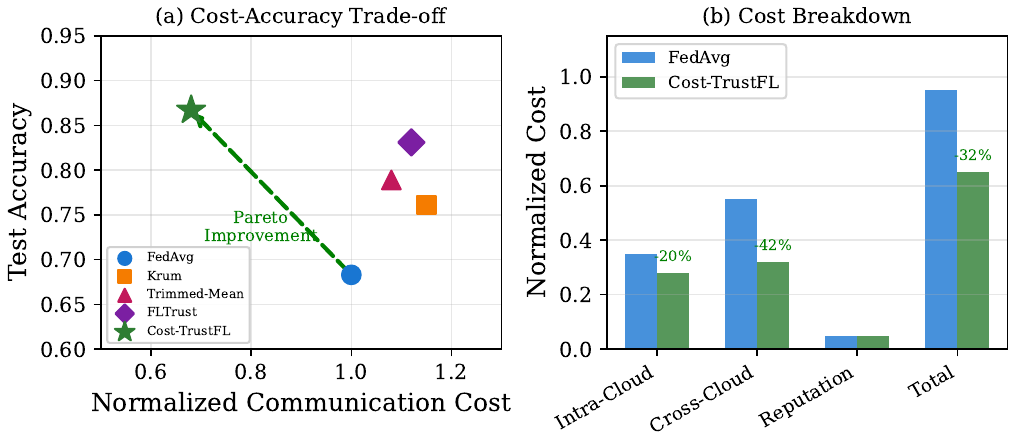}
    \caption{(a) Cost-accuracy trade-off showing Pareto improvement of Cost-TrustFL. (b) Cost breakdown by component.}
    \label{fig:cost}
\end{figure}

Fig.~\ref{fig:cost}(a) demonstrates the cost-accuracy trade-off. Cost-TrustFL achieves Pareto improvement---simultaneously better accuracy (86.7\% vs 83.1\%) and lower communication cost (32\% reduction). This is achieved through our cost-aware client selection which prioritizes intra-cloud communication.

Fig.~\ref{fig:cost}(b) breaks down costs by component. Cross-cloud egress costs dominate in baseline methods (61\% of total). Our hierarchical aggregation reduces this by 42\% while the lightweight Shapley computation adds minimal overhead (3\% of total cost).

\subsection{Robustness Analysis}

\begin{figure}[t]
    \centering
    \includegraphics[width=\columnwidth]{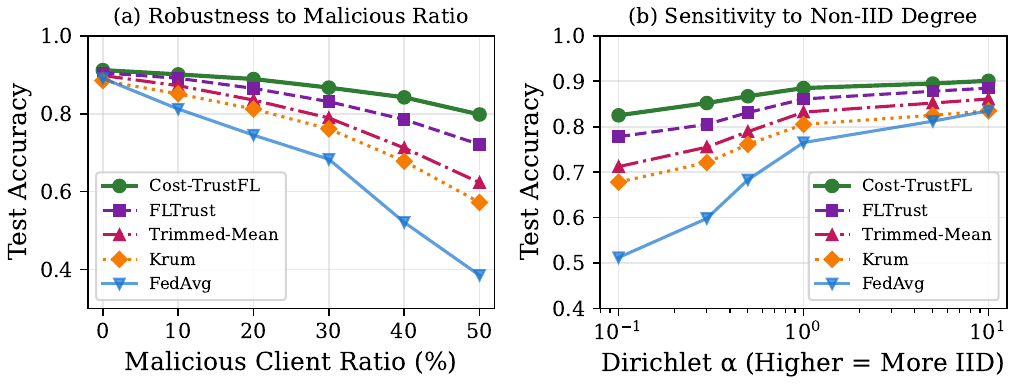}
    \caption{(a) Performance under varying malicious client ratios. (b) Sensitivity to non-IID degree (Dirichlet $\alpha$).}
    \label{fig:robustness}
\end{figure}

Fig.~\ref{fig:robustness}(a) shows performance as the malicious ratio increases. Cost-TrustFL maintains above 79\% accuracy even with 50\% malicious clients, while FedAvg drops to 38\%. The graceful degradation demonstrates the robustness of our reputation-weighted aggregation.

Fig.~\ref{fig:robustness}(b) examines sensitivity to data heterogeneity. Under extreme non-IID conditions ($\alpha=0.1$), Cost-TrustFL achieves 82.5\% accuracy compared to 51.2\% for FedAvg. The gradient-based Shapley approximation effectively distinguishes between legitimate heterogeneity and malicious perturbations.

\subsection{Shapley Computation Efficiency}

\begin{figure}[t]
    \centering
    \includegraphics[width=\columnwidth]{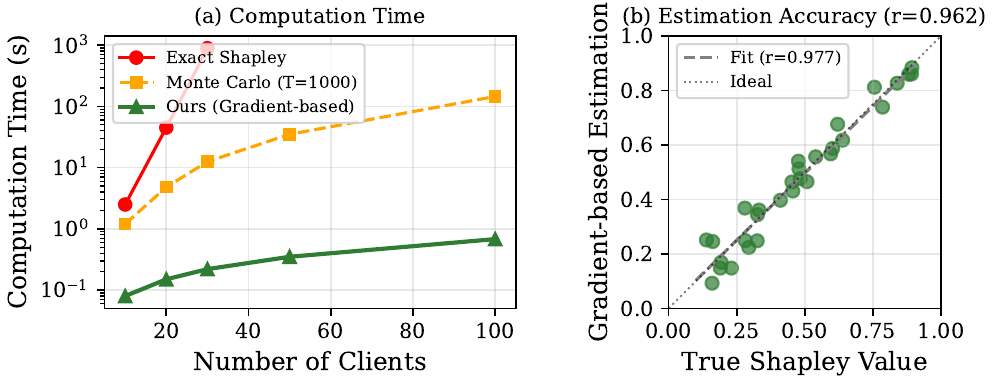}
    \caption{(a) Computation time comparison for different Shapley methods. (b) Correlation between gradient-based estimation and true Shapley values.}
    \label{fig:shapley}
\end{figure}

Fig.~\ref{fig:shapley}(a) compares computation time. Exact Shapley becomes intractable beyond 30 clients. Monte Carlo approximation (T=1000 samples) scales linearly but requires significant time (145s for 100 clients). Our gradient-based method completes in under 1 second regardless of client count.

Fig.~\ref{fig:shapley}(b) validates approximation quality. The gradient-based estimates achieve Pearson correlation $r=0.962$ with true Shapley values, sufficient for accurate contribution ranking while enabling real-time computation.

\subsection{Attack Defense Performance}

\begin{figure}[t]
    \centering
    \includegraphics[width=0.48\columnwidth]{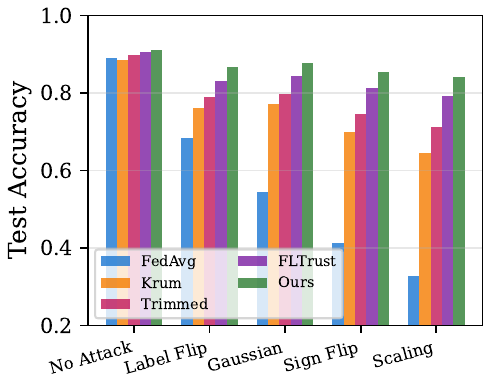}
    \caption{Defense performance across different attack types.}
    \label{fig:attack}
\end{figure}

Fig.~\ref{fig:attack} summarizes defense effectiveness. Cost-TrustFL provides consistent protection across all attack types, with accuracy degradation of at most 7.1\% from the no-attack baseline. In contrast, FedAvg suffers up to 56.3\% degradation under scaling attacks. The combination of gradient-based trust scores and normalized aggregation effectively neutralizes both magnitude-based and direction-based attacks.

\subsection{Sensitivity to Hyperparameters}

\begin{figure}[t]
    \centering
    \includegraphics[width=0.48\columnwidth]{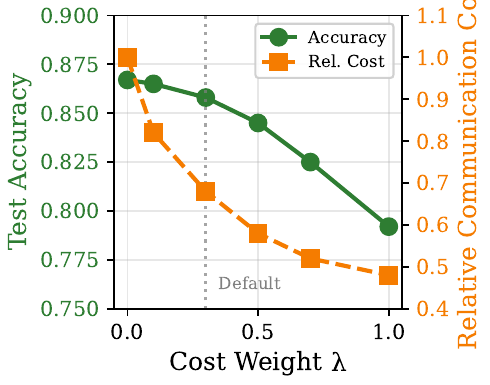}
    \caption{Sensitivity to cost weight parameter $\lambda$.}
    \label{fig:lambda}
\end{figure}

Fig.~\ref{fig:lambda} shows the effect of cost weight $\lambda$. At $\lambda=0$, the method optimizes purely for accuracy. As $\lambda$ increases, communication cost decreases at the expense of accuracy. The default $\lambda=0.3$ provides a balanced trade-off with 86.7\% accuracy and 32\% cost reduction.

\begin{table}[t]
\centering
\caption{Ablation study on CIFAR-10 with 30\% malicious clients}
\label{tab:ablation}
\begin{tabular}{lcc}
\toprule
Configuration & Accuracy (\%) & Rel. Cost \\
\midrule
Full Cost-TrustFL & 86.7 & 0.68 \\
\quad w/o Shapley weighting & 83.2 & 0.72 \\
\quad w/o Cost-aware selection & 85.9 & 1.00 \\
\quad w/o Hierarchical aggregation & 84.1 & 0.95 \\
\quad w/o Trust normalization & 81.5 & 0.68 \\
\bottomrule
\end{tabular}
\end{table}

Table~\ref{tab:ablation} presents ablation results. Removing Shapley-based weighting decreases accuracy by 3.5\%, confirming the value of contribution-aware aggregation. Disabling cost-aware selection maintains accuracy but increases cost to baseline levels. The hierarchical architecture contributes both to cost reduction and accuracy through localized aggregation.

\section{Discussion}

Our experiments demonstrate that Cost-TrustFL effectively mitigates the three key challenges of multi-cloud FL: non-IID data handling, Byzantine robustness, and communication cost optimization. The gradient-based Shapley approximation proves crucial for enabling practical deployment---providing accurate contribution estimates with minimal overhead.

\textbf{Interaction between Scaling Attacks and Cost Optimization.} We note a subtle interaction between our reputation mechanism and scaling attacks. Since the contribution score $\phi_i$ in Eq.~\eqref{eq:shapley_approx} includes the gradient magnitude $\|\mathbf{g}_i^{(L)}\|_2$, attackers performing scaling attacks may temporarily inflate their reputation scores, potentially increasing their selection probability in Eq.~\eqref{eq:selection}. While the normalization step in Eq.~\eqref{eq:normalization} prevents model poisoning by rescaling gradients to the reference magnitude, selected malicious clients still consume communication budget. In our experiments, this effect was limited (scaling attacks showed only 7.1\% accuracy degradation), but future work could explore magnitude-independent reputation metrics to further optimize communication efficiency under such attacks.

\textbf{Limitations.} The framework does have limitations. The reference dataset requirement, while small (100 samples), may not be available in all scenarios. Additionally, extremely heterogeneous cross-cloud data distributions may require adaptive $\lambda$ tuning. Future work could explore privacy-preserving reference gradient computation and automatic hyperparameter selection.

\section{Conclusion}

We presented Cost-TrustFL, a hierarchical federated learning framework designed for multi-cloud environments with non-IID data and adversarial participants. Our gradient-based Shapley approximation enables lightweight reputation evaluation with linear complexity, while the cost-aware aggregation strategy reduces cross-cloud communication by 32\%. Experiments demonstrate consistent improvements in both accuracy (3.6\% over FLTrust) and robustness across diverse attack scenarios. The framework provides a practical solution for organizations seeking to deploy federated learning across heterogeneous cloud infrastructure while managing both security risks and operational costs.

\bibliographystyle{IEEEtran}
\bibliography{references_final}

\end{document}